\documentclass[letterpaper, 10 pt, conference]{ieeeconf}  
\IEEEoverridecommandlockouts
\usepackage{graphics} 
\usepackage{amssymb}  

\usepackage{subcaption}
\usepackage{wrapfig}
\usepackage{mathtools, cuted}

\usepackage{pdflscape} 
\usepackage{lipsum}
\usepackage{lineno}
 \usepackage{amsmath}
 \usepackage{tikz}
\usetikzlibrary{shapes.geometric, arrows}

\usepackage{url}
\usepackage[font=normalsize]{subcaption}
\usepackage{multirow,array,diagbox}
\usepackage{tabularx}
\usepackage{adjustbox}
\usepackage{lscape}
\usepackage{xcolor,colortbl}
\usepackage{tikz}

\usepackage{multicol,multirow}
\usepackage{pifont}
\usepackage{arydshln}
\usepackage{booktabs}
\usepackage{makecell}
\usepackage{threeparttable} 
\usepackage{algorithm} 
\usepackage{algorithmic}

\usepackage{xcolor}

\definecolor{Gray}{gray}{0.85}
\definecolor{LightCyan}{rgb}{0.88,1,1}

\newcolumntype{a}{>{\columncolor{Gray}}c}
\newcolumntype{b}{>{\columncolor{white}}c}

\title{\LARGE \bf
Cross-Embodiment Affordance Transfer for Robotic Manipulation
}

\author{Sawera Yaseen$^{*1}$, Serge Thill$^{1}$, Rekha Raja$^{1}$
\thanks{$^{1}$School of Artificial Intelligence, Donders Centre for Cognition, Department of Human-Centred Intelligent Systems, Radboud University, Nijmegen, The Netherlands.}
\thanks{*Email: sawera.yaseen@donders.ru.nl}}

\begin{document}

\maketitle
\thispagestyle{empty}
\pagestyle{empty}

\begin{abstract}
Robotic manipulation systems commonly learn object affordances from human-annotated datasets, which reflect interaction possibilities for a human hand rather than the physical constraints of a specific robot. For example, a surface region of an object identified as graspable by a human annotator may be geometrically unreachable or mechanically infeasible for a  robot, yet this mismatch is rarely addressed explicitly. To tackle this challenge, we proposed a framework that translates human affordance annotations into robot-executable affordance predictions conditioned on the physical properties of the robot. The network predicts which surface regions of object are physically executable by the robot, given a 3D point cloud of an object and a descriptor encoding the robot's geometry and constraints. The model was trained using human-annotated affordances together with robotic execution outcomes from multiple sampled approach trials in simulation, where execution outcomes determined which regions are executable for the robot. The evaluation was performed on a household objects dataset using two parallel-jaw gripper configurations on a UR5e robot arm. The framework outperformed a model that does not receive information about the robot. The experimental results demonstrated that affordance prediction must be conditioned on the robot and that human annotations alone are insufficient for learning robot-executable affordance representations.

\end{abstract}
\maketitle

 \noindent \textbf{\textit{Keywords}} - Affordance learning, cross-embodiment, embodiment-conditioned perception, robotic manipulation.

\section{Introduction}
Robots operating in the real world must be able to determine not only what objects are, but also how they can be interacted with to accomplish a task. In ecological psychology, J. J. Gibson \cite{Gibson1979} introduced the term ``affordance'' in a discussion of how humans can directly perceive such action possibilities in the environment. In robotics, the ecological aspect of direct perception, that is, without internal modelling and representation, is usually lost; instead, affordance-based representations link visual perception to manipulation, enabling systems to identify where and how to interact with objects.
\cite{Yamanobe2017,ardon2021building}. Recent advances in visual affordance prediction have demonstrated promising results in grasp detection, manipulation planning, and functional object understanding using RGB and RGB-D perception and deep learning methods \cite{10341553,7139369}.
These methods construct affordance representations from existing affordance datasets which provide valuable information about how humans perceive possible interactions with objects. In particular, human-annotated datasets \cite{10222906,7139369} 
identify object regions associated with functional interactions providing a foundation for learning where and how objects can be used. However, the assumptions underlying these annotations introduce a fundamental challenge that limits how far such representations can be transferred to robots.
\begin{figure}
    \centering
    \includegraphics[width=\columnwidth]{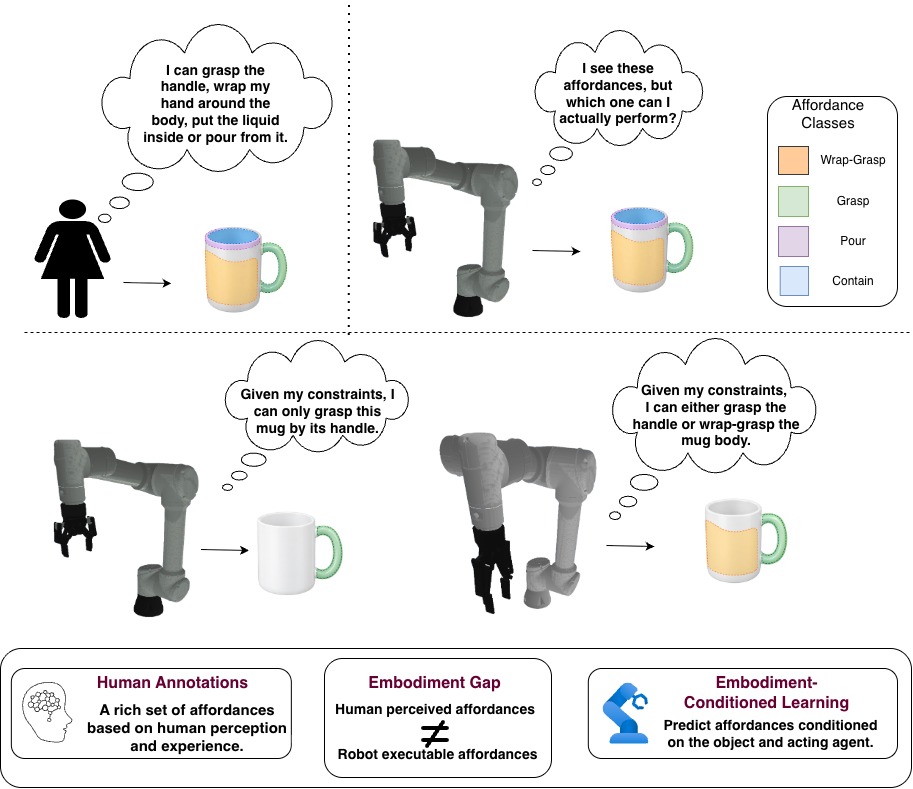}
    \caption{Human-to-robot affordance gap. Human-interpreted affordance regions do not necessarily correspond to robot-executable regions, as executability depends on the physical capabilities of the acting embodiment.}
    \label{fig:intro}
\end{figure}
Human-annotated affordance datasets describe potential interactions from a human perspective without grounding them in the execution capabilities of a particular robot. This limitation is well recognised. Myers et al.~\cite{7139369} provide region-level annotations derived from human observation, and subsequent work on this dataset notes that gripper and robot specifications restrict prediction generalisability in real-world settings~\cite{liu2024affordancekeypoint}. Cavalli et al.~\cite{cavalli2019affordanceprediction} question the validity of human-labelled annotations when the acting agent is not a human hand, and Kemmeren et al.~\cite{kemmeren2024affordances} observe that such datasets fail to address how robot embodiment affects interaction feasibility. The underlying issue is that affordances are relational, arising from the interaction between an object and a specific agent. A dexterous human hand can approach and grasp from many locations and orientations, whereas a parallel-jaw gripper with limited aperture and restricted motion may access only a subset of these regions. Transferring human annotations directly to a robot therefore assumes that human-perceived affordance is equally actionable by that robot, an assumption that breaks down across embodiments with different morphologies and physical capabilities.

To address this limitation, we investigated the human-to-robot affordance gap (Fig. \ref{fig:intro}) and proposed an embodiment-conditioned affordance learning framework. We formulated affordance prediction as a problem that jointly reasons over visual object geometry and robot feasibility. As a source of human interpreted affordance knowledge, we use the UMD Part Affordance Dataset~\cite{7139369}, a dataset providing affordance labels across a range of categories for common household and tool objects, supplemented by robotic execution trials that provide embodiment-specific supervision. Object geometry is represented as point clouds and predictions are conditioned on a descriptor encoding the robot's geometry and constraints. Given an object point cloud and an embodiment descriptor, the model predicts which surface regions are both geometrically associated with an affordance and physically executable for that embodiment, focusing on grasp and wrap-grasp as the primary interaction modes of parallel-jaw grippers. Which parts of human-derived affordance knowledge transfer and which require adaptation is itself embodiment-dependent. The framework makes this analysis tractable for a specific robot without manual re-annotation, using existing human annotations as a common representation of object-interaction knowledge while relatively small amounts of robotic interaction data adapt this knowledge to a given embodiment. We evaluate on the Yale-CMU-Berkeley (YCB) dataset~\cite{calli2015benchmarking}, investigating whether human-interpreted affordance regions transfer directly to robots, how embodiment conditioning changes predicted actionable regions, and whether robotic execution data improves generalisation across embodiments.

The main contributions of this work are threefold. First, we characterised the gap between human-interpreted affordances and robot-executable affordances, showing through experiments across multiple embodiments that affordance regions derived from human annotations cannot be assumed directly transferable to a robotic platform.
Second, we proposed an embodiment-conditioned affordance learning framework that combines human-annotated affordance knowledge with robotic execution outcomes to model the relationship between object geometry and embodiment-specific interaction feasibility. Third, we
experimentally demonstrated that the learned robot-executable affordances generalise to object categories unseen during training, while retaining embodiment-specific feasibility.

\section{Related work}

\subsection{Visual Affordance Detection and Segmentation}
Early affordance detection methods framed the problem as pixel-level segmentation over RGB or RGB-D images. Myers et al.~\cite{7139369} introduced the UMD dataset with per-part annotations across 105 tools and 7 affordance classes, and Do et al.~\cite{8460902} proposed AffordanceNet for joint object detection and affordance mask prediction. These 2D approaches established visual affordance learning from human-annotated regions but offer limited geometric information for manipulation and remain viewpoint-sensitive.

The shift to 3D point clouds addressed this. Deng et al.~\cite{9577339} established point-level affordance segmentation as a benchmark with 3D AffordanceNet, Nguyen et al.~\cite{10341553} extended prediction to open-vocabulary settings, He et al.~\cite{He2026} integrated 2D semantic priors with 3D reasoning for scene-level segmentation, and Vu et al.~\cite{Vu2026} scaled affordance learning to complex indoor scenes via keypoint matching. Despite these advances, all treat affordances as object properties under a fixed embodiment, without modelling how the acting agent's physical capabilities influence executability.

\subsection{Embodiment-Conditioned and Cross-Embodiment Affordance Learning}

Affordance predictions depend on multiple factors such as object geometry, task semantics, physical object properties, and the morphology and kinematics of the acting agent \cite{ardon2021building}.
These conditioning axes have been explored largely in isolation. Geometric conditioning through point cloud encoders has been addressed via multi-scale local feature aggregation ~\cite{10610593}.
Task-semantic conditioning through natural language has enabled models to activate different output regions depending on intended manipulation \cite{10341553}.
Bahl et al. \cite{Bahl_2023_CVPR} used affordances extracted from human videos as a versatile conditioning substrate for downstream robotic manipulation. Borja-Diaz et al. \cite{9811889} showed that affordance representations learned from unstructured play data can condition sample-efficient policy learning without task-specific annotation. Conditioning on agent embodiment has been explored primarily outside the affordance domain. Chen et al. \cite{Chen2018} embedded hardware descriptors into policy networks to enable transfer across manipulators with different morphologies. Attarian
et al. \cite{pmlr-v229-attarian23a} addressed multi embodiment grasping through shared geometric representations across gripper geometries. Schiavi et al. \cite{10160747} conditioned affordance predictions on the agent's configuration state for closed-loop articulated object interaction, though within a single fixed embodiment.
Across these conditioning axes, embodiment properties have not been used to condition per-point affordance segmentation. Geometric and language-conditioned methods carry no representation of the acting manipulator, and embodiment-encoding methods operate at the policy level, focusing on control transfer rather than modelling affordances as embodiment-dependent interactions.

\subsection{Human-Interpreted vs. Robot-Executable Affordances}
Although substantial progress has been made in  affordance prediction, task-conditioned learning, and embodiment-aware manipulation, existing approaches largely assume that affordances can be transferred across embodiments with limited adaptation. Furthermore, publicly available datasets rarely combine human-interpreted affordance annotations with robotic manipulation outcomes across different embodiments, particularly unsuccessful interactions. Consequently, the relationship between human affordance interpretation, embodiment, physical execution remains insufficiently understood. In this work, we address this gap by distinguishing between human-interpreted affordances and robot-executable affordances. We treat human annotations as a source of object-interaction knowledge and use successful and failed robotic executions to learn how this knowledge depends on the physical capabilities of the acting embodiment. This formulation enables affordance knowledge to be adapted to different robotic embodiments while retaining its transferability to novel objects.

\section{Method}
\subsection{Problem Formulation}
Given an object point cloud and a robot embodiment descriptor, the objective is to predict the affordance regions of the object that are executable by the robot.
Let $\mathcal{P} \in \mathbb{R}^{N \times 3}$ be the object point cloud expressed in robot base frame, where $N$ is the number of points. Human affordance annotations from the UMD dataset \cite{7139369} are represented as a label vector:
\begin{equation}
    \mathbf{a}^H = \{a_i^H\}_{i=1}^{N},
    \quad a_i^H \in \{0, 1, 2\}
\end{equation}
where $a_i^H = 0$ indicates grasp, $a_i^H = 1$ indicates wrap-grasp, and $a_i^H = 2$ indicates background. 

Let $\mathbf{e} \in \mathbb{R}^{D}$ be a descriptor encoding the physical configuration of the robot, where $D$ is the dimensionality of the embodiment representation.  Formally the affordance prediction function is defined as:
\begin{equation}
    G_\theta : (\mathcal{P},\; \mathbf{e})
    \;\longrightarrow\; \hat{\mathbf{y}},
    \qquad \hat{y}_i \in \{0, 1, 2\}
\end{equation}
where $\hat{y}_i \in \{0, 1\}$ indicates a robot-feasible affordance and $\hat{y}_i = 2$ indicates background. Points predicted as
background include both geometrically unannotated regions and affordance regions that are physically infeasible for the given robot at the observed object pose, which together define the
human-to-robot affordance gap. Human affordance annotations $\mathbf{a}^H$ provide training supervision only and are not available to the model at inference time.

\subsection{Multi-Embodiment Data Generation}

\subsubsection{Human-Interpreted Data}
The affordance annotations are sourced from the UMD RGB-D Part Affordance dataset \cite{7139369}, which provides per-point affordance labels. Here we only retain grasp and wrap-grasp labels as target classes and all remaining classes are mapped to background, yielding the human annotation vector $\mathbf{a}^H \in \{0, 1, 2\}^N$. These annotations capture geometrically valid interaction regions on each object surface, independent of embodiment, and serve as the geometric prior from which embodiment-conditioned labels are derived.

\subsubsection{Robot Interaction Data Generation}
We developed a simulation pipeline to generate the robotic interaction data. The pipeline uses PyBullet as the physics engine with a combined UR5e robot and Robotiq 2F-85 and Robotiq 2F-140 grippers. Each object instance is placed at 200 sampled poses on a tabletop workspace, modelled as a free rigid body subject to gravity and contact dynamics.
Candidate contact points are obtained by reprojecting UMD affordance annotations onto the object point cloud using Iterative Closest Point (ICP) registration \cite{Besl1992}, which
aligns the UMD source point cloud to the target object point cloud by minimising the mean point-to-point distance between corresponding surfaces. This yields the annotated surface set $\mathcal{A}_H$ per scene.

For each candidate point $p_i \in \mathcal{A}_H$, five approach poses are sampled and the robot attempts the
corresponding manipulation primitive from each. Inverse kinematics is solved and collision checking is performed at each waypoint. Each trial receives one of four verdicts based on execution outcome. A trial is marked \textsc{pass} if the gripper achieves contact at the affordance region, closes without collision, and lifts the object stably. It is marked \textsc{partial}
if the gripper reaches the target but fails to close, \textsc{mismatch} if no valid IK solution is found, and \textsc{miss} if no feasible approach trajectory exists.

A candidate point $p_i$ receives its affordance class label $k \in \{0, 1\}$ if at least one of the five approach trials is marked \textsc{pass}, and is assigned class~$2$ otherwise. Points in $\mathcal{A}_H \setminus \mathcal{A}_R(\mathbf{e})$ are thereby labelled background, encoding the human-to-robot affordance gap in the label space and providing per-point supervision
for for learning embodiment-dependent robot-executable affordances.

\subsubsection{Embodiment Representation}
An embodiment descriptor $\mathbf{e} \in \mathbb{R}^{D}$ encodes the physical properties of a robot that determine which interactions it
can execute. To represent factors that can influence execution feasibility, the descriptor captures three aspects of the robot embodiment. The reachable workspace defines which object poses the robot can reach. The end-effector geometry determines whether the gripper can physically close around an object region. The joint limits constrain which approach poses are kinematically feasible. Omitting any of these leaves the model with an incomplete picture of what the robot can physically execute. Any descriptor that encodes these three aspects for a given robot can be used with the proposed framework, and the dimensionality $D$ varies with the robot configuration.
For the UR5e robot used in this work, the descriptor is instantiated as a 21-dimensional vector partitioned into three groups:
\begin{equation}
    \mathbf{e} =
    \begin{bmatrix}
        \mathbf{w}_{4 \times 1} \\
        \mathbf{g}_{5 \times 1} \\
        \mathbf{q}_{12 \times 1}
    \end{bmatrix}
    \in \mathbb{R}^{21}
\end{equation}
The kinematic workspace vector $\mathbf{w} \in \mathbb{R}^{4}$ encodes outer horizontal reach, inner dead-zone radius, and minimum and maximum end-effector height. The end-effector geometry vector $\mathbf{g} \in \mathbb{R}^{5}$ encodes maximum and minimum jaw opening, finger length, finger pad contact width, and lateral palm reach offset. The joint-limit vector $\mathbf{q} \in \mathbb{R}^{12}$ encodes the lower and upper bounds of each of the six UR5e joints, interleaved per joint:
\begin{equation}
    \mathbf{q} =
    \begin{bmatrix}
        q_0^{\min} & q_0^{\max} &
        \cdots &
        q_j^{\min} & q_j^{\max}
    \end{bmatrix}^{\top}
\end{equation}
where $q_j^{\min}$ and $q_j^{\max}$ are the lower and upper rotation limits of joint $j$ in radians, for $j \in \{0,\dots,5\}$ corresponding to the base, shoulder, elbow, and three wrist
joints of the UR5e arm. The 12-dimensional joint-limit block reflects the six-DOF configuration of the UR5e. A robot with a
different number of joints produces a descriptor of different dimensionality. The embodiment descriptor is used to condition affordance learning and enable cross-embodiment transfer. The cross-embodiment transfer is achieved by substituting $\mathbf{e}$ with the descriptor of the target embodiment, requiring no architectural modification. The descriptor carries no learned parameters and is fixed per robot configuration making the representation directly extensible to new robots. 

\subsection{Proposed Network Architecture}
\begin{figure*}
    \centering
    \includegraphics[width=0.8\textwidth]{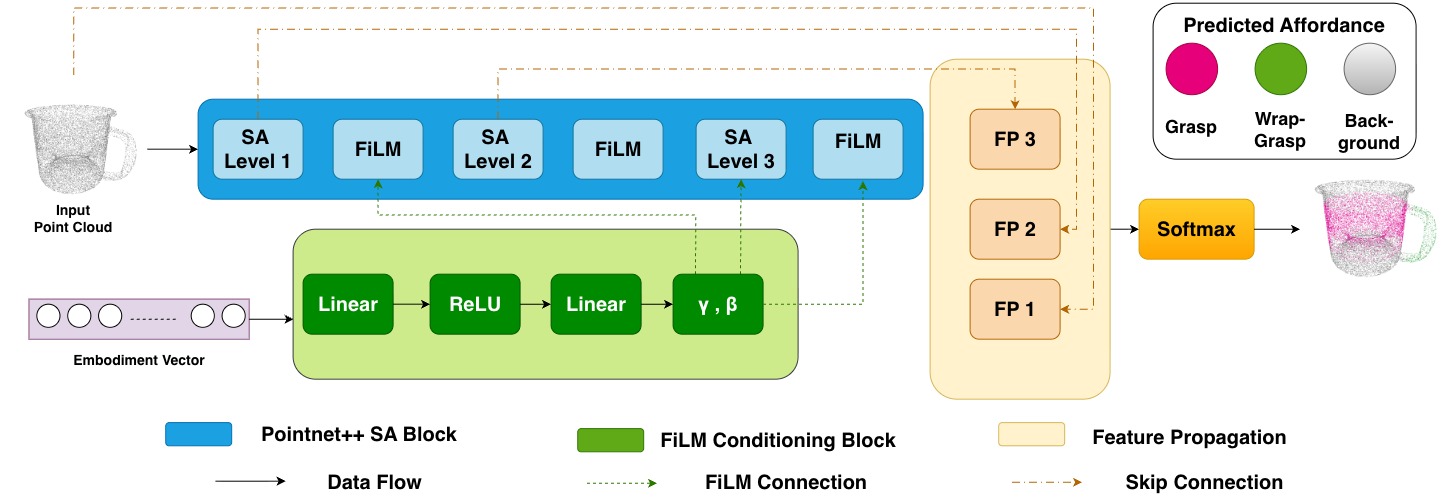}
    \caption{FiLM-conditioned PointNet++ for affordance segmentation. The model takes a 2048-point cloud and a 21-dimensional embodiment descriptor $\mathbf{e}$. A per-point segmentation head produces logits over three classes: grasp, wrap-grasp, and background.}
    \label{fig:archi}
\end{figure*}

We propose a hierarchical point segmentation network that conditions geometric feature learning on the acting embodiment through stage-wise affine modulation. The embodiment descriptor $\mathbf{e}$ is injected at each Set Abstraction (SA) stage, allowing the geometric representations to adapt to embodiment-specific physical capabilities at different levels of spatial abstraction. 
The architecture comprises a PointNet++ encoder with multi-scale grouping (MSG), stage-wise Feature-wise Linear Modulation (FiLM) conditioning blocks, and a per-point classification head, as illustrated in Fig.~\ref{fig:archi}.

\subsubsection{PointNet++ Encoder}
We use PointNet++ \cite{qi2017pointnetplusplus} with MSG as the backbone. Three SA stages progressively downsample the point cloud while aggregating local geometric features across multiple query radii at each level:
\begin{itemize}
    \item SA$_1$: $N \!\rightarrow\! 512$ centroids,
          radii $\{0.1, 0.2, 0.4\}$\,m,
          output 320 channels
    \item SA$_2$: $512 \!\rightarrow\! 128$ centroids,
          radii $\{0.2, 0.4, 0.8\}$\,m,
          output 640 channels
    \item SA$_3$: global aggregation, output 1024 channels
\end{itemize}
Three Feature Propagation (FP) layers upsample back to the original $N$-point resolution using distance-weighted interpolation and skip connections from the corresponding SA stages. These skip connections preserve fine-grained geometric information required for point-level affordance prediction.

\subsubsection{FiLM Conditioning}
At each SA level $l \in \{1, 2, 3\}$, a dedicated two-layer MLP
$g_\theta^{(l)} : \mathbb{R}^{D} \rightarrow \mathbb{R}^{C_l}
\times \mathbb{R}^{C_l}$ where $\theta$ denotes the learnable parameters, maps
the embodiment descriptor to per-channel affine parameters:
\begin{equation}
    \boldsymbol{\gamma}^{(l)},\;
    \boldsymbol{\beta}^{(l)}
    \;=\; g_\theta^{(l)}(\mathbf{e}),
    \qquad
    \boldsymbol{\gamma}^{(l)},\,\boldsymbol{\beta}^{(l)}
    \in \mathbb{R}^{C_l}
\end{equation}
where $\boldsymbol{\gamma}^{(l)}$ is the per-channel scale vector and $\boldsymbol{\beta}^{(l)}$ is the per-channel shift vector at level $l$. The  $C_l \in \{320, 640, 1024\}$ is the feature dimensionality at level $l$. The SA output features are then modulated via
Feature-wise Linear Modulation (FiLM) \cite{perez2018film}:
\begin{equation}
    \tilde{\mathbf{h}}_i^{(l)}
    \;=\; \boldsymbol{\gamma}^{(l)} \odot \mathbf{h}_i^{(l)}
    \;+\; \boldsymbol{\beta}^{(l)},
    \qquad \forall\; i \in \{1,\dots,N_l\}
    \label{eq:film}
\end{equation}
where $N_l$ is the number of centroids at level $l$ and $\odot$
denotes elementwise multiplication. The generators
$\{g_\theta^{(l)}\}_{l=1}^{3}$ are instantiated independently
per stage with non-shared parameters, transforming the conditioning
signal separately at each feature dimensionality $C_l$. The
modulated features $\tilde{\mathbf{h}}^{(l)}$ are forwarded to the
FP layers via skip connections, propagating embodiment influence
through to the full-resolution per-point representations.
 
\subsubsection{Classification Head}The final FP output is processed by two shared 1D convolutional layers with 256 and 128 channels, respectively. Each layer is followed by batch normalization, ReLU activation, and dropout. A final linear projection maps the resulting per-point features to three output logits, corresponding to the grasp class, wrap-grasp class, and background. During inference, a softmax operation converts these logits into per-point class probabilities.

\subsection{Training Objective}The network is trained end-to-end using the embodiment-conditioned labels obtained from human affordance annotations and robotic execution outcomes. Points outside the human-interpreted affordance regions are assigned the background class. For candidate affordance points within $\mathcal{A}_H$, successful execution retains the corresponding affordance class $k\in \{0,1\}$, whereas unsuccessful execution is assigned the background class, indicating that the human-interpreted affordance is not executable for the evaluated embodiment. We use weighted cross-entropy to account for the class imbalance between background and affordance points:
\begin{equation}
    \mathcal{L}
    = -\frac{1}{N}\sum_{i=1}^{N}
      w_{y_i} \log \hat{p}_{i,\,y_i}
\end{equation}
where $y_i \in \{0,1,2\}$ is the ground-truth label at point $i$, $\hat{p}_{i,k}$ is the predicted probability of class $k$, and $w_{y_i}$ is the weight assigned to the ground-truth class. Class weights are defined as:
\begin{equation}
    w_k = \left(
    \frac{\max_j N_j}{N_k}
    \right)^{\!\!1/3}
\end{equation}
with $N_k$ the total number of training points of class $k$.

\section{Experiments}
\subsection{Experimental Setup}
\begin{figure}
    \centering
    \includegraphics[scale=0.4]{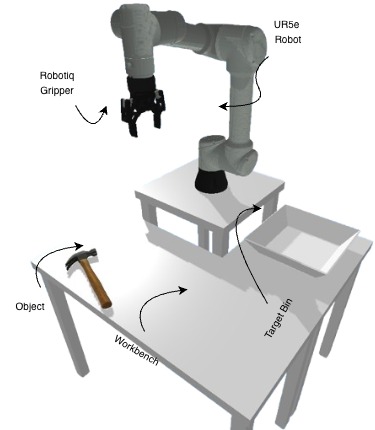}
    \caption{Experimental setup in the PyBullet simulation. It consists of a UR5e manipulator with interchangeable Robotiq 2F-85 and 2F-140 grippers on a tabletop workspace.}
    \label{fig:workspace}
\end{figure}

We evaluate the proposed cross-embodiment affordance learning framework on a simulated UR5e robotic manipulation platform equipped with an end-effector mounted RGB-D camera (Fig. \ref{fig:workspace}). The system is designed to study affordance transfer between human-interpreted affordance annotations and robotic execution under varying embodiment configurations. We consider two robot embodiments: (i) a UR5e manipulator equipped with a Robotiq 2F-85 gripper, and (ii) a UR5e manipulator equipped with a Robotiq 2F-140 gripper. Human annotations from the UMD dataset provide the
human-interpreted affordance supervision. These configurations enable studying affordance transfer under significant morphological and interaction differences. The evaluation considers manipulation tasks including grasping and wrap grasping for different objects. For each task, the model predicts affordance regions corresponding to feasible interactions for the target embodiment, which are subsequently used to execute the associated robotic action.

\subsection{Baseline}
The proposed embodiment-conditioned model is compared against a no-embodiment baseline in which the embodiment descriptor is not provided as input. This ablation tests whether object geometry alone
is sufficient for affordance prediction.

\subsection{Implementation Details.}
The model is trained end-to-end using the Adam optimizer ($\beta_1 = 0.9$, $\beta_2 = 0.999$) with an initial learning rate of \textit{$5 \times 10^{-4}$} and weight decay \textit{$1 \times 10^{-4}$}. The learning rate schedule applies linear warmup over \textit{10} epochs, then cosine annealing to $1 \times 10^{-6}$. Training runs for a maximum of \textit{150} epochs with early stopping at a patience of 40 epochs and a batch size of \textit{8}.  All experiments are conducted on a single  \textit{NVIDIA Quadro RTX 6000} on an  \textit{Intel Xeon Gold 5218} server. 

\subsection{Evaluation Metrics}
We evaluate point-wise affordance prediction performance using precision, recall, F1-score, and Intersection-over-Union (IoU). These metrics are computed separately for each affordance class $k \in \{0, 1\}$. Let $\text{TP}_k$, $\text{FP}_k$, and $\text{FN}_k$ denote the true positives, false positives, and false negatives for class $k$. The per-class metrics are:
\begin{equation}
    \text{Prec}_k = \frac{\text{TP}_k}{\text{TP}_k + \text{FP}_k},
    \qquad
    \text{Rec}_k  = \frac{\text{TP}_k}{\text{TP}_k + \text{FN}_k}
\end{equation}
\begin{equation}
    \text{F1}_k   = \frac{2\,\text{Prec}_k \cdot \text{Rec}_k}
                         {\text{Prec}_k + \text{Rec}_k},
    \qquad
    \text{IoU}_k  = \frac{\text{TP}_k}
                         {\text{TP}_k + \text{FP}_k + \text{FN}_k}
\end{equation}
Mean IoU (mIoU) is computed as the macro-average over the $K$ affordance classes:

\begin{equation}
    \text{mIoU} = \frac{1}{K}\sum_{k=1}^{K} \text{IoU}_k
\end{equation}

mIoU is the primary metric for comparing across experiments. The gain of the embodiment-conditioned
model over the no-embodiment baseline is reported as:
\begin{equation}
    \Delta\text{mIoU} =
    \text{mIoU}_{\text{EC}} - \text{mIoU}_{\text{NE}}
\end{equation}
where $\text{mIoU}_{\text{EC}}$ and
$\text{mIoU}_{\text{NE}}$ denote the mIoU of the
embodiment-conditioned model and the no-embodiment
baseline respectively.

\section{Results}

\subsection{Affordance Segmentation Accuracy}
Table~\ref{tab:combined-ablation} compares the embodiment conditioned model against the no-embodiment baseline across both affordance classes. Embodiment conditioning yields consistent improvements across every reported metric, with a macro
mIoU gain of 1.18 points. The consistency of
this gain across all metrics and both affordance classes demonstrates that the embodiment descriptor carries information complementary to object geometry and that stage-wise FiLM conditioning successfully propagates this signal through the geometric feature hierarchy. These results validated the central claims of this work. Human-annotated affordance regions were not directly transferable to robot execution without adaptation, and explicitly conditioning
on the robot's physical configuration reduced this gap, producing more accurate predictions of physically executable regions than reasoning from object geometry alone.
The two affordance classes exhibit structurally different prediction behaviour. Wrap-grasp achieves near-perfect precision in both conditions, indicating the model rarely predicts wrap-grasp outside the true executable region. Grasp exhibits high recall with lower precision, reflecting the broader spatial distribution of graspable regions across the object surface. Embodiment conditioning improves both classes, with the larger absolute gain on grasp, where manipulator-specific constraints such as
jaw clearance and approach angle have more influence on which regions are physically reachable.

\begin{table}
\centering
\caption{Affordance segmentation performance and
embodiment-conditioning ablation. $\Delta$ denotes the mIoU
improvement of the embodiment-conditioned model over the
no-embodiment baseline.}
\label{tab:combined-ablation}
\setlength{\tabcolsep}{3.5pt}
\begin{tabular}{llcccc}
\toprule
\textbf{Method} & \textbf{Affordance} &
\textbf{Prec.} & \textbf{Rec.} & \textbf{F1} & \textbf{IoU} \\
\midrule
Embodiment-conditioned & Grasp
& 44.31 & 89.66 & 59.31 & 42.15 \\
& Wrap-Grasp
& 99.92 & 65.16 & 78.88 & 65.12 \\
& \textbf{Macro Avg.}
& \textbf{72.11} & \textbf{77.41} & \textbf{69.09} & \textbf{53.64} \\
\midrule
No-embodiment baseline & Grasp
& 43.32 & 86.61 & 57.76 & 40.60 \\
& Wrap-Grasp
& 99.83 & 64.39 & 78.29 & 64.32 \\
& \textbf{Macro Avg.}
& \textbf{71.58} & \textbf{75.50} & \textbf{68.02} & \textbf{52.46} \\
\midrule
\multicolumn{2}{l}{\textbf{$\Delta$ IoU}}
& -- & -- & -- & \textbf{+1.18} \\
\bottomrule
\end{tabular}
\vspace{-6pt}
\end{table}

\subsection{Cross-Gripper Transfer}
Table~\ref{tab:aperture-ablation} presents cross-gripper transfer results, where each model is trained on one gripper configuration and evaluated against ground-truth labels from the other. The results reveal a pronounced and interpretable asymmetry in transfer direction.

A model trained on the Robotiq 2F-140 transfers
competitively to 2F-85 evaluation. The wider aperture
exposes the model to successful wrap-grasp executions
across a broader range of object geometries, producing
richer affordance supervision. Adapting to the more
constrained 2F-85 at evaluation requires predicting a
subset of the learned feasibility landscape, a
structurally easier problem since the model has already observed the relevant geometry and associated it with feasible interactions during training.
The reverse direction is substantially harder. A model
trained on 2F-85 data showed limited transfer to 2F-140
evaluation, with a wrap-grasp IoU of 26.63\% compared
to 63.26\% for the 2F-140-trained model. The 2F-85 physically cannot wrap-grasp most objects due to its narrower aperture, producing fewer successful wrap-grasp trials during training. The model consequently acquires no representation of feasible wrap-grasp regions and fails systematically when evaluated against ground truth where wrap-grasp is widely executable. This failure is consistent with a supervision gap caused by the embodiment gap rather than solely by model capacity. A manipulator that cannot execute an interaction cannot provide the training signal needed to learn it.

The two embodiment configurations in this experiment
differ only in maximum aperture of the gripper, while sharing identical arm kinematics, workspace parameters, and joint limits.
Despite this limited inter-embodiment variation, the
impact on affordance feasibility is substantial.
Embodiments with larger kinematic differences, such
as varying reach, joint limits, or fundamentally
different morphologies, would expose a wider
embodiment gap and produce more pronounced
differentiation in affordance predictions. The full
21-dimensional embodiment descriptor is designed to accommodate such variation, and evaluation across more diverse embodiments remains an important direction for future work.
 
\begin{table}
  \caption{Cross gripper transfer performance. The grippers used are Robotiq 2F-85 and 2F-140. Each model is trained on one gripper and evaluated on another. $\Delta$ is computed as 2F-140\, $-$ 2F-85\,.}
  \label{tab:aperture-ablation}
  \centering
  \setlength{\tabcolsep}{4pt}
  \begin{tabular}{ll cccc}
    \hline
    \textbf{Affordance} & \textbf{Embodiment}
      & \textbf{Prec.} & \textbf{Rec.}
      & \textbf{F1}   & \textbf{IoU}\\
    \hline
    \multirow{3}{*}{Grasp}
      & 2F-85\,   & 44.72 & 90.53 & 59.87 & 42.72 \\
      & 2F-140\,  & 43.71 & 87.26 & 58.25 & 41.09 \\
      & $\Delta$ & \textbf{-1.01} & \textbf{-3.27} & \textbf{-1.62} & \textbf{-1.63} \\
    \hline
    \multirow{3}{*}{Wrap-Grasp}
      & 2F-85\,   & 99.45 & 26.67 & 42.06 & 26.63 \\
      & 2F-140\,  & 99.80 & 63.34 & 77.49 & 63.26 \\
      & $\Delta$ & \textbf{+0.35} & \textbf{+36.67} & \textbf{+35.43} & \textbf{+36.63} \\
    \hline
    \multirow{3}{*}{\textbf{Aggregated}}
      & 2F-85\,   & 72.08 & 58.60 & 50.96 & 34.68 \\
      & 2F-140\,  & 71.76 & 75.30 & 67.87 & 52.17 \\
      & $\Delta$ & \textbf{-0.32} & \textbf{+16.70} & \textbf{+16.91} & \textbf{+17.49} \\
    \hline
  \end{tabular}
\end{table}

\subsection{Qualitative Analysis}
\begin{figure*}
    \centering
    \setlength{\tabcolsep}{2pt}

    \begin{tabular}{c c c c c c c c}
        &
        \multicolumn{5}{c}{\textbf{Unseen Objects}}
        &
        \multicolumn{2}{c}{\textbf{Seen Objects}}
        \\[1mm]

        \rotatebox{90}{\footnotesize RGB}
        &
        \includegraphics[width=0.125\textwidth,height=2.0cm]{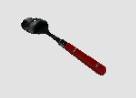}
        &
        \includegraphics[width=0.125\textwidth,height=2.0cm]{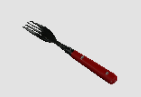}
        &
        \includegraphics[width=0.125\textwidth,height=2.0cm]{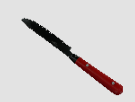}
        &
        \includegraphics[width=0.125\textwidth,height=2.0cm]{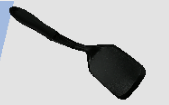}
        &
        \includegraphics[width=0.125\textwidth,height=2.0cm]{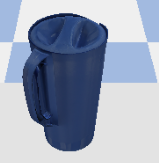}
        &
        \includegraphics[width=0.125\textwidth,height=2.0cm]{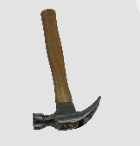}
        &
        \includegraphics[width=0.125\textwidth,height=2.0cm]{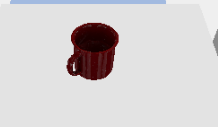}
        \\[2mm]

        \rotatebox{90}{\footnotesize Depth}
        &
        \includegraphics[width=0.125\textwidth,height=2.0cm]{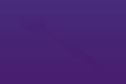}
        &
        \includegraphics[width=0.125\textwidth,height=2.0cm]{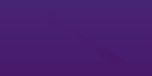}
        &
        \includegraphics[width=0.125\textwidth,height=2.0cm]{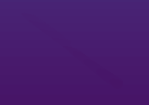}
        &
        \includegraphics[width=0.125\textwidth,height=2.0cm]{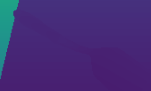}
        &
        \includegraphics[width=0.125\textwidth,height=2.0cm]{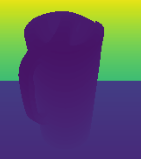}
        &
        \includegraphics[width=0.125\textwidth,height=2.0cm]{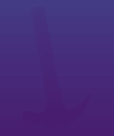}
        &
        \includegraphics[width=0.125\textwidth,height=2.0cm]{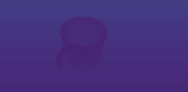}
        \\[1mm]

        \rotatebox{90}{\footnotesize Point Cloud}
        &
        \includegraphics[width=0.125\textwidth,height=2.0cm]{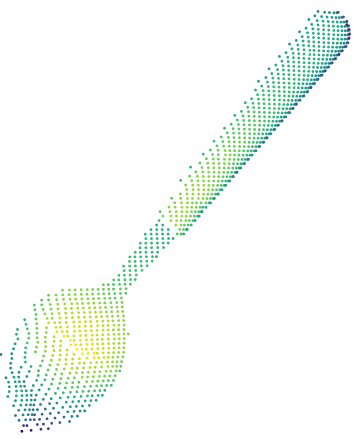}
        &
        \includegraphics[width=0.125\textwidth,height=2.0cm]{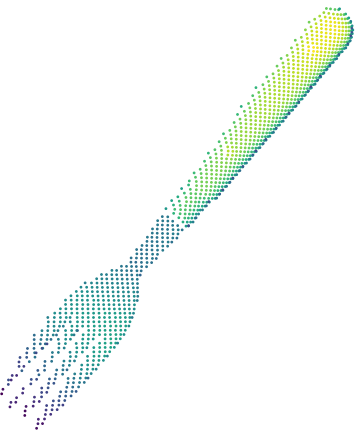}
        &
        \includegraphics[width=0.125\textwidth,height=2.0cm]{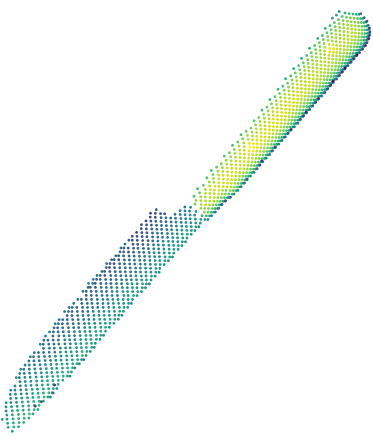}
        &
        \includegraphics[width=0.125\textwidth,height=2.0cm]{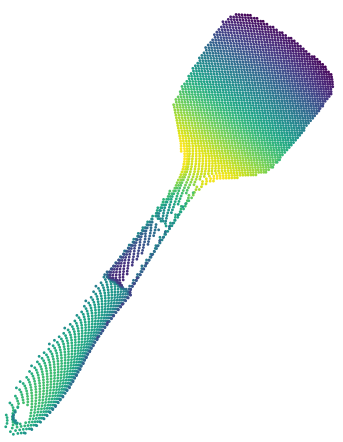}
        &
        \includegraphics[width=0.125\textwidth,height=2.0cm]{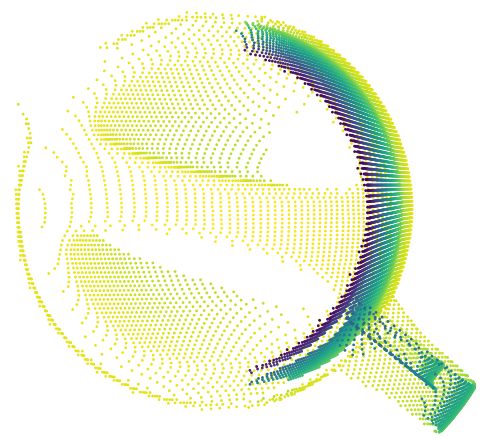}
        &
        \includegraphics[width=0.125\textwidth,height=2.0cm]{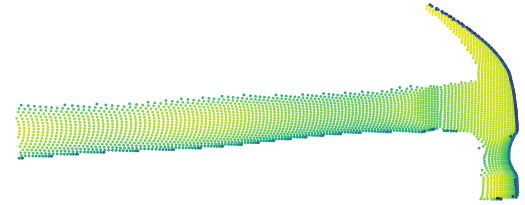}
        &
        \includegraphics[width=0.125\textwidth,height=2.0cm]{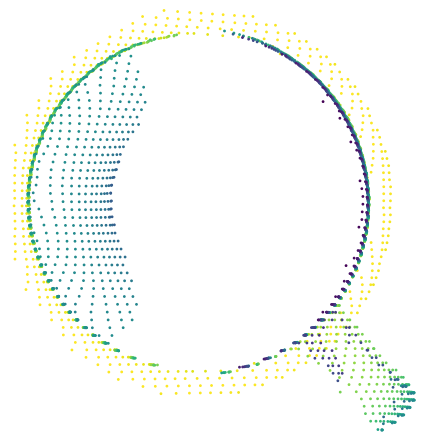}
        \\[1mm]

        \rotatebox{90}{\footnotesize Affordance}
        &
        \includegraphics[width=0.125\textwidth,height=2.0cm]{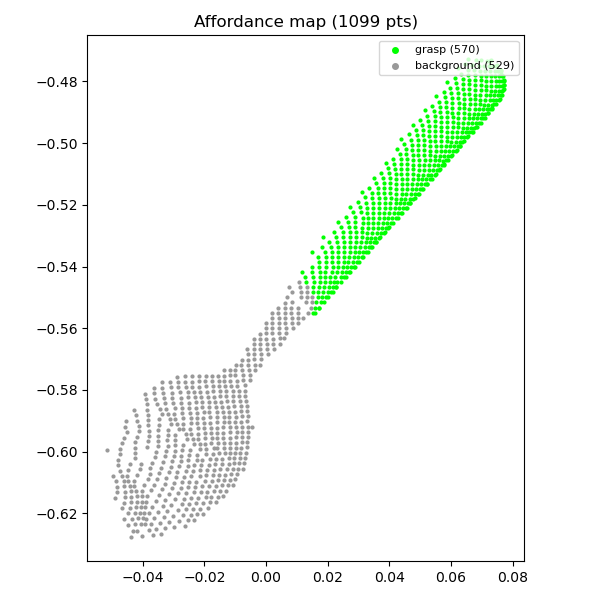}
        &
        \includegraphics[width=0.125\textwidth,height=2.0cm]{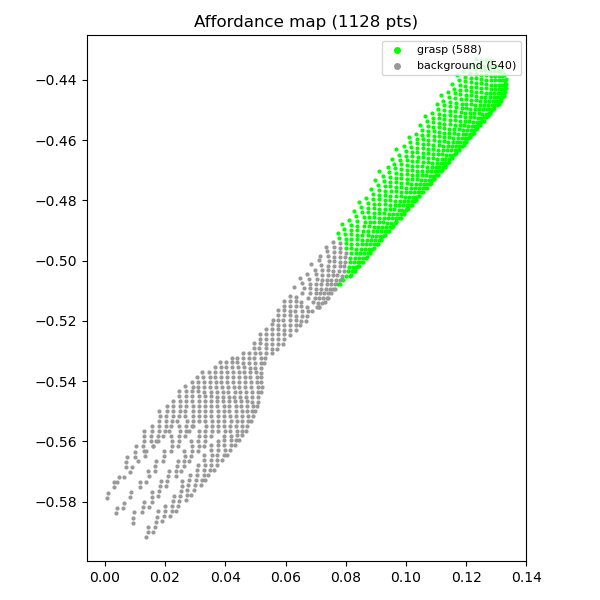}
        &
        \includegraphics[width=0.125\textwidth,height=2.0cm]{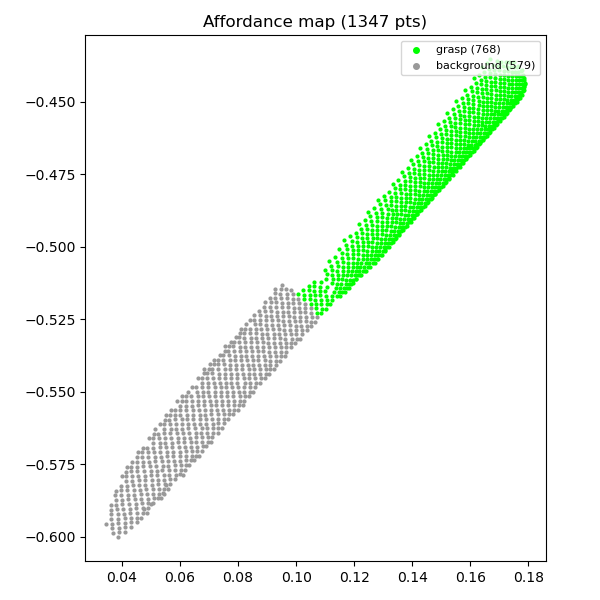}
        &
        \includegraphics[width=0.125\textwidth,height=2.0cm]{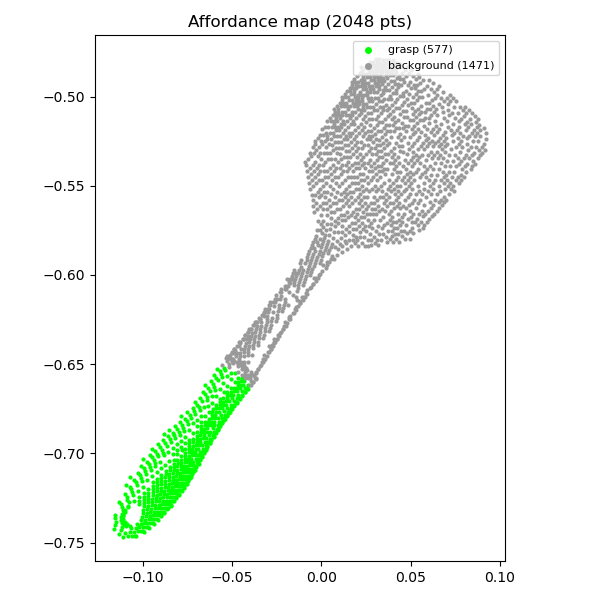}
        &
        \includegraphics[width=0.125\textwidth,height=2.0cm]{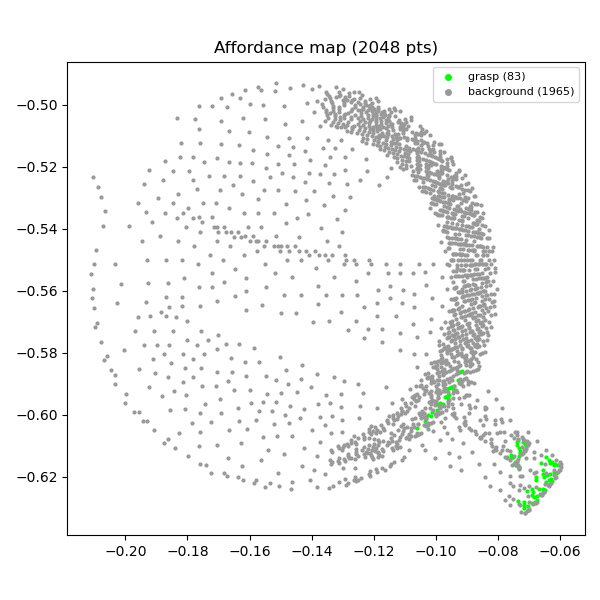}
        &
        \includegraphics[width=0.125\textwidth,height=2.0cm]{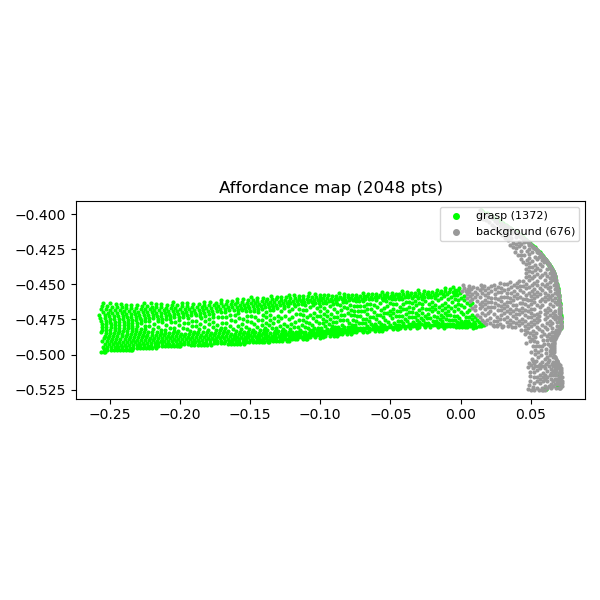}
        &
        \includegraphics[width=0.125\textwidth,height=2.0cm]{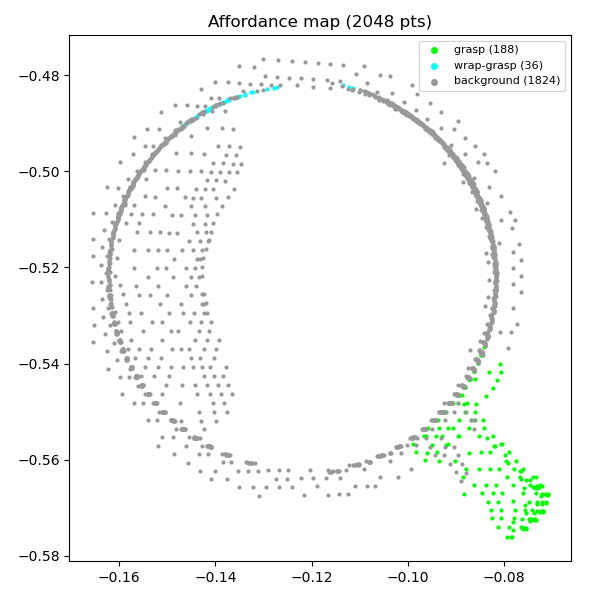}
        \\

    \end{tabular}

    \caption{Qualitative results across seen and unseen objects. Each column corresponds to an object, while rows show the RGB observation, depth representation, reconstructed point cloud, and predicted affordance map.}
    \label{fig:qualitative-results}
\end{figure*}

Figure~\ref{fig:qualitative-results} shows per-point affordance
predictions across seen and unseen object categories.
On unseen objects, predicted grasp regions consistently
corresponded to geometrically graspable regions such
as tool handles and mug handles, demonstrating
cross-category generalisation without retraining.
On seen objects, predictions remained consistent with the executable regions for the Robotiq
2F-85 gripper. 

\subsection{Physical Geometry Grounds the Embodiment Gap}
\begin{figure}
    \centering
    \setlength{\tabcolsep}{1pt}

    \begin{tabular}{ccccc}
        \includegraphics[width=0.19\columnwidth,height=1.6cm]{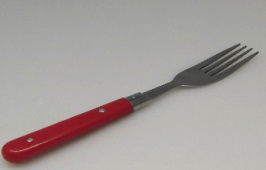} &
        \includegraphics[width=0.19\columnwidth,height=1.6cm]{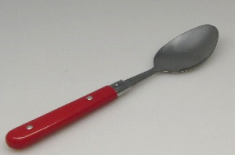} &
        \includegraphics[width=0.19\columnwidth,height=1.6cm]{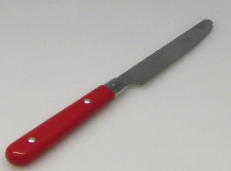} &
        \includegraphics[width=0.19\columnwidth,height=1.6cm]{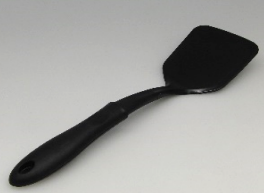} &
        \includegraphics[width=0.19\columnwidth,height=1.6cm]{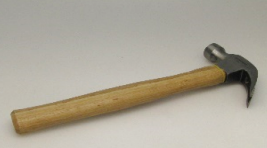}
        \\[0.5mm]

        \includegraphics[width=0.19\columnwidth,height=1.6cm]{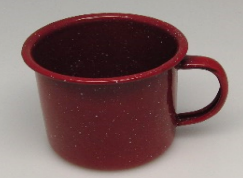} &
        \includegraphics[width=0.19\columnwidth,height=1.6cm]{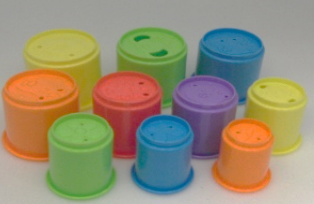} &
        \includegraphics[width=0.19\columnwidth,height=1.6cm]{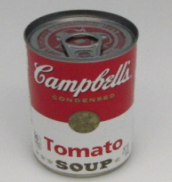} &
        \includegraphics[width=0.19\columnwidth,height=1.6cm]{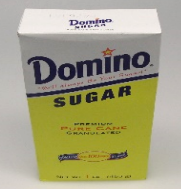} &
        \includegraphics[width=0.19\columnwidth,height=1.6cm]{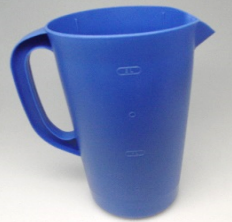}
    \end{tabular}

    \caption{Representative YCB objects used for evaluating predicted affordances across both gripper configurations. Corresponding physical feasibility analysis is given in Tab.~\ref{tab:object_affordance}.}
    \label{fig:objects}
\end{figure}

Table~\ref{tab:object_affordance} grounds the
numerical results in physical geometry,
characterising which affordances are valid for each
gripper across the object set shown in Fig.~\ref{fig:objects}. Several objects that geometrically afford wrap-grasp, such as mugs and
cans with rounded cylindrical bodies, are infeasible
for the 2F-85 because their body diameter exceeds
the 85\,mm aperture, while the same objects are
executable for the 2F-140. Flat objects do not afford wrap-grasp for either gripper, confirming that the interaction is
constrained by both object geometry and embodiment
simultaneously.

This analysis makes explicit what the numerical
results imply. The wrap-grasp performance gap in
Table~\ref{tab:aperture-ablation} is not a
consequence of model capacity but follows directly from the embodiment gap. A
human annotator correctly identifies the mug body
as wrap-graspable, yet the 2F-85, constrained by
its aperture, cannot realise that affordance while
the 2F-140 can. Table~\ref{tab:object_affordance}
provides the physical grounding that connects the
learned representations to manipulator reality,
confirming that embodiment-conditioned affordance
prediction is not a modelling choice but a physical
necessity.

\begin{table}
\caption{Physical affordance feasibility across two gripper configurations and object geometries. Grasp: feasible if the graspable cross-section is less than the gripper maximum span. Wrap-grasp: feasible if the gripper span exceeds the object body diameter and the object is cylindrical or rounded. Flat objects are not amenable to wrap-grasp.}
\label{tab:object_affordance}

\centering
\renewcommand{\arraystretch}{1.2}

\resizebox{\columnwidth}{!}{%
\begin{tabular}{l c c c c c}
\toprule
\multirow{2}{*}{\textbf{Object}} &
\multirow{2}{*}{\textbf{Dimensions (mm)}} &
\multicolumn{2}{c}{\textbf{Robotiq 2F-85}} &
\multicolumn{2}{c}{\textbf{Robotiq 2F-140}} \\
\cmidrule(lr){3-4}
\cmidrule(lr){5-6}
& & \textbf{Grasp} & \textbf{Wrap-Grasp} 
& \textbf{Grasp} & \textbf{Wrap-Grasp} \\
\midrule

Spoon
& 186 $\times$ 83 $\times$ 21
& \checkmark & \textemdash
& \checkmark & \textemdash \\

Fork
& 198 $\times$ 27 $\times$ 16
& \checkmark & \textemdash
& \checkmark & \textemdash \\

Knife
& 215 $\times$ 23 $\times$ 21
& \checkmark & \textemdash
& \checkmark & \textemdash \\

Spatula
& 306 $\times$ 125 $\times$ 33
& \checkmark & \textemdash
& \checkmark & \textemdash \\

Hammer
& 333 $\times$ 182 $\times$ 33
& \checkmark & \textemdash
& \checkmark & \textemdash \\

Mug
& 117 $\times$ 93 $\times$ 81
& \texttimes & \texttimes
& \checkmark & \checkmark \\

Mug
& 100 $\times$ 81 $\times$ 72
& \checkmark & \checkmark
& \checkmark & \checkmark \\

Mug
& 100 $\times$ 92 $\times$ 58
& \texttimes & \texttimes
& \checkmark & \checkmark \\

Master Chef Can
& 140 $\times$ 103 $\times$ 102
& \texttimes & \texttimes
& \checkmark & \checkmark \\

Sugar Box
& 176 $\times$ 94 $\times$ 50
& \checkmark & \textemdash
& \checkmark & \textemdash \\

Cup
& 57 $\times$ 57 $\times$ 62
& \checkmark & \textemdash
& \checkmark & \textemdash \\

Tomato Soup Can
& 102 $\times$ 68 $\times$ 68
& \checkmark & \checkmark
& \checkmark & \checkmark \\

Pitcher Base
& 242 $\times$ 149 $\times$ 145
& \texttimes & \texttimes
& \texttimes & \texttimes \\

\bottomrule
\end{tabular}%
}

\vspace{2pt}

\footnotesize
$\checkmark$ feasible; $\times$ infeasible; -- not applicable.

\end{table}

\section{Discussion}

\subsection{Key Findings}
The experimental results addressed the three questions raised in the introduction. The same object induced different interaction regions depending on the agent's morphology and action capabilities. Regions correctly identified as wrap-graspable by a human annotator were infeasible for a parallel-jaw gripper whose aperture did not accommodate the object diameter, yet executable for a wider configuration. This was a systematic property of the human-to-robot affordance gap, not an edge case, arguing against treating human-annotated affordances as intrinsic object properties directly transferable across embodiments. Explicitly modelling embodiment information improved this transfer, and jointly incorporating object geometry with embodiment produced more discriminative affordance representations than object-centric baselines, particularly where human hand interactions did not map directly to robot gripper kinematics.

Conditioning affordance prediction on embodiment allowed the model to align human-annotated affordances with robot-executable interactions. Supervision derived from robotic execution outcomes, including failed attempts, provided the signal needed to distinguish which annotated regions were executable for a given robot. The degree of generalisation across embodiments depended on transfer direction, revealing that the availability of embodiment-specific supervision was itself a function of what the robot could physically execute.

\subsection{Limitations and Future Work}
The evaluation covers two gripper configurations sharing arm kinematics and differing only in aperture. This is sufficient to demonstrate the embodiment gap's existence, but produces small inter-embodiment variation in the descriptor. Configurations with larger kinematic differences such as arm reach, joint limits, or end-effector morphology would expose a wider affordance gap and stronger prediction differentiation. The object and task scope is similarly constrained to YCB objects in simulation across two affordance classes. Broader object sets, additional affordance types, and real-world validation would strengthen generality. Conditioning gains are expected to scale with morphological and kinematic dissimilarity, as the descriptor carries a stronger discriminative signal. Validating this across diverse robotic platforms is the primary direction for future work.

\section{Conclusion}

In this work, we studied robotic affordances as embodiment-dependent interactions rather than intrinsic object properties, and proposed a cross-embodiment affordance learning framework that jointly modelled object geometry and
manipulator configuration. A central finding was that existing human-annotated affordance datasets, while valuable as a source of object-interaction knowledge, were not directly transferable to robots with different bodies.
The proposed framework increased the utility
of such datasets by adapting their knowledge
to a specific embodiment through robotic
execution data, without requiring manual
re-annotation for each new platform. Experiments on a UR5e robot with two parallel-jaw
gripper configurations demonstrated that
conditioning affordance prediction on the
robot's physical properties improved prediction
accuracy and reduced the gap between
human-annotated affordances and robot-executable
regions. The learned representations generalised
to object categories unseen during training,
and transfer across gripper configurations
showed that the availability of
embodiment-specific supervision was a function
of what the robot could physically execute. Explicitly modelling embodiment proved necessary
for learning physically grounded affordance
representations from human-annotated data.
Existing affordance datasets can serve as a
scalable foundation for cross-embodiment
manipulation learning when paired with targeted
robotic interaction data, and this work opens
a path toward broader generalisation across
robot platforms without the cost of platform-specific annotation.

\bibliographystyle{IEEEtran}
\bibliography{IEEEexample}
 
\end{document}